%% file: root.tex
\documentclass[letterpaper, 10 pt, conference]{ieeeconf}

\IEEEoverridecommandlockouts
\usepackage{graphicx}
\usepackage{amsmath}
\usepackage{amssymb}
\usepackage{xspace}
\usepackage{multirow}
\usepackage{booktabs}
\usepackage{subcaption}
\usepackage{fvextra}
\usepackage{xcolor}
\usepackage{threeparttable}
\usepackage{makecell}
\usepackage{capt-of}
\usepackage{dirtytalk}
\usepackage{url}
\usepackage[hidelinks]{hyperref}

\definecolor{promptbg}{gray}{0.97}
\definecolor{promptframe}{gray}{0.80}
\definecolor{linkblue}{RGB}{56,95,145}
\DefineVerbatimEnvironment{prompt}{Verbatim}{
  fontsize=\small,
  frame=single,
  rulecolor=\color{promptframe},
  framesep=3mm,
  baselinestretch=1.0,
  breaklines=true
}
\DefineVerbatimEnvironment{smallprompt}{Verbatim}{
  fontsize=\scriptsize,
  frame=single,
  rulecolor=\color{promptframe},
  framesep=3mm,
  baselinestretch=1.0,
  breaklines=true
}

\newcommand{\ourvla}{Point-VLA\xspace}

\title{\LARGE \bf
Point What You Mean: Visually Grounded Instruction Policy
}

\author{
\textbf{Hang Yu}$^{1,3\ast}$ \quad
\textbf{Juntu Zhao}$^{2,3\ast}$ \quad
\textbf{Yufeng Liu}$^{2,3}$ \quad
\textbf{Kaiyu Li}$^{3}$ \quad
\textbf{Cheng Ma}$^{3}$ \quad
\textbf{Di Zhang}$^{1}$ \\
\textbf{Yingdong Hu}$^{4}$ \quad
\textbf{Guang Chen}$^{1}$ \quad
\textbf{Junyuan Xie}$^{3}$ \quad
\textbf{Junliang Guo}$^{3\ddagger}$ \quad
\textbf{Junqiao Zhao}$^{1\dagger}$ \quad
\textbf{Yang Gao}$^{3,4\dagger}$ \\
$^{1}$Tongji University \quad
$^{2}$Shanghai Jiao Tong University \quad
$^{3}$Spirit AI \quad
$^{4}$Tsinghua University
}

\begin{document}

\twocolumn[{
    \begin{@twocolumnfalse}
    \maketitle
    \renewcommand{\thefootnote}{\fnsymbol{footnote}}
    \footnotetext[1]{Equal contribution. Work done during the internship at Spirit AI.}
    \footnotetext[2]{Corresponding authors. $^{\ddagger}$Project leader.}
    \thispagestyle{empty}
    \pagestyle{empty}
    \begin{center}
        \includegraphics[width=0.9\textwidth]{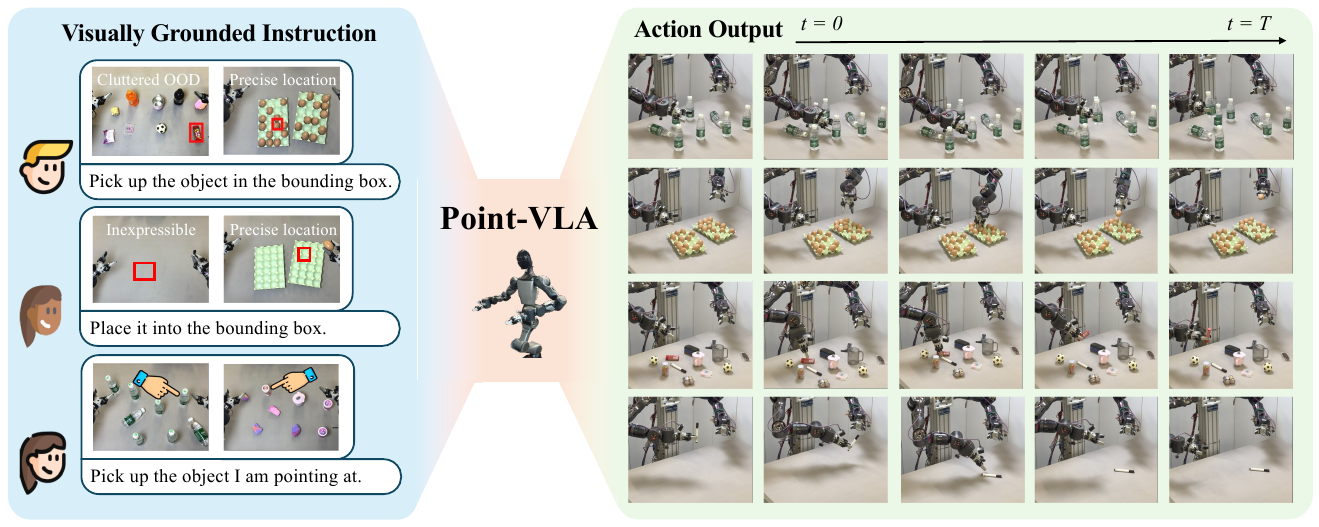}
        \vspace{-5pt}
        \captionof{figure}{We introduce \ourvla, which resolves the inherent limitations of text-only instructions in precise target referring, e.g., referring objects in clutter, handling unseen OOD objects, or placing on plain tabletop without reference point. By overlaying bounding boxes on images, \ourvla provides explicit pixel-level cues that enable accurate and unambiguous referring in real-world manipulation.}
        \label{fig:teaser}
    \end{center}
    \vspace{5pt}
    \end{@twocolumnfalse}
}]

\begin{abstract}

Vision–Language–Action (VLA) models align vision and language with embodied control, but their object referring ability remains limited when relying solely on text prompt, especially in cluttered or out-of-distribution (OOD) scenes.
In this study, we introduce the Point-VLA, a plug-and-play policy that augments language instructions with explicit visual cues (e.g., bounding boxes) to resolve referential ambiguity and enable precise, object-level grounding.
To efficiently scale visually grounded datasets, we further develop an automatic data annotation pipeline requiring minimal human effort.
We evaluate Point-VLA on diverse real-world referring tasks and observe consistently stronger performance than text-only instruction VLAs, particularly in cluttered or unseen-object scenarios, with robust generalization.
These results demonstrate that Point-VLA effectively resolves object referring ambiguity through pixel-level visual grounding, achieving more generalizable embodied control.

\vspace{3pt}
\noindent\textbf{Project page:} \href{https://yuhang-harry.github.io/Point-VLA/}{\textcolor{linkblue}{\nolinkurl{https://yuhang-harry.github.io/Point-VLA/}}}

\end{abstract}

\section{INTRODUCTION}
\label{sec:intro}

Embodied intelligence has achieved substantial progress in recent years~\cite{chi2023diffusion,zhao2023learning,arxiv2023_openxembodiment_rtx,Black2025RealTimeEO,liu2026learningnativecontinuationaction}. Benefiting from advanced large-scale multimodal foundation models, Vision-Language-Action (VLA) systems now exhibit strong text-conditioned action execution capabilities~\cite{bjorck2025gr00t,liu2024rdt,arxiv2025_ren_pi05_openworld,arxiv2024_ren_pi0_flow_vla}.

Despite remarkable progress, Vision–Language–Action (VLA) models remain fundamentally constrained by the inherent information bottleneck of the text modality\cite{zhao2025sayingunsaidrevealinghidden}.
Textual instructions alone cannot precisely describe all referential situations, many real-world objects or spatial relations are difficult to specify through language alone~\cite{Krahmer2012ComputationalGO,Tellex2020RobotsTU,Li2016SpatialRA,Guadarrama2013GroundingSR,Liu2021PerspectiveCorrectedSR,Paul2017GroundingAS}.
While visual grounding has been extensively studied in Vision-Language Models (VLMs) for understanding and description tasks, VLA grounding presents fundamentally different challenges: VLMs output text descriptions, whereas VLAs must generate precise motor actions for physical manipulation. This difference creates unique technical requirements for spatial precision, temporal consistency, and action-level grounding that existing VLM techniques do not address.
As a result, VLA models are unable to leverage visual cues such as pointing, gaze, or gesture to resolve referential ambiguity, often failing to establish robust linguistic grounding in cluttered or unfamiliar scenarios and thus limiting their generalization capabilities.
Specifically, we identify two core challenges:

\input{figures/method/inference}
\vspace{-3pt}

\begin{itemize}
    \item \textbf{Inexpressible references.}
    Language alone cannot precisely specify irregular or amorphous objects (e.g., a lump of clay), exact spatial targets on plain surfaces, or specific items in cluttered scenes. For instance, describing ``the bottle to the right of the leftmost bottles, in the middle of the desk'' among eight identical bottles remains ambiguous even for state-of-the-art VLMs.

    \item \textbf{Limited generalization.}
    Text-based VLA models struggle when instructions involve complex spatial references or novel object categories. Even when VLMs can localize objects with 60-70\% accuracy, text-only VLAs achieve only 25\% manipulation success, revealing a fundamental language-action alignment gap.
\end{itemize}

We introduce \ourvla, which overlays bounding boxes on camera images to provide explicit pixel-level referents (Figure~\ref{fig:teaser}), enabling precise visual grounding alongside standard multi-view observations.
We co-train the model on both textual and visual grounded instructions to obtain a single unified policy that can operate in either pure-text or visually grounded modes.

Beyond the core architecture, we adopt an automatic annotation strategy by utilizing Multi-modal Large Language Models~(MLLMs) to propose target bounding boxes from demonstration videos.
We apply two grounding-aware augmentations—random translation and localized appearance perturbation—to decouple grounding from absolute coordinates and improve generalization.
These choices keep visual grounding supervision easy to obtain while enabling \ourvla's ability to generalize to unseen objects and spatial configurations.

Across six referent-sensitive manipulation tasks, Point-VLA achieves 92.5\% average success, with +35 and +75 point improvements on unseen-object and egg-slot picking respectively, while remaining fully compatible with text-only instructions.

Here we summarize our contributions:
\begin{itemize}
    \item We propose \ourvla, a VLA model that augments linguistic instructions with explicit visual grounding, resolving referential ambiguity and enabling precise object and location specification in cluttered and previously unseen environments.
    \item \ourvla is a unified policy that supports either instruction mode while preserving strong text-following behavior, thanks to co-training on both text-only and visually grounded instructions.
    \item We provide a scalable data construction pipeline that leverages pretrained MLLMs to automatically generate visually grounded supervision from existing trajectories, reducing annotation cost and supporting seamless integration with prior datasets.
\end{itemize}

\section{RELATED WORK}

\input{figures/method/training}

\paragraph{Generalist VLA and spatial understanding.}
Large-scale Vision--Language--Action (VLA) models advance generalist robotic control by unifying perception, language, and action in a single policy~\cite{Zhao2025CoTVLAVC,Lin2025OneTwoVLAAU,Wang2025VQVLAIV}.
Early systems such as \textsc{RT-1} and \textsc{RT-2} demonstrate that scaling data and model capacity yields broad task coverage, while \textsc{Octo}, \textsc{OpenVLA}, \textsc{DROID}, BridgeData, LIBERO, and related open benchmarks provide standardized datasets and baselines for cross-embodiment training~\cite{arxiv2022_brohan_rt1,arxiv2023_brohan_rt2,arxiv2023_openxembodiment_rtx,arxiv2024_octo_opensource_policy,corl2024_kim_openvla,khazatsky2024droid,walke2023bridgedata,Jones2025BeyondSF,Liu2023LIBEROBK,Dasari2024TheIF}.
Recent \textsc{$\pi_0$} and \textsc{$\pi_{0.5}$} further improve robustness through flow-matched action generation in open-world settings~\cite{arxiv2024_ren_pi0_flow_vla,arxiv2025_ren_pi05_openworld}.
To strengthen spatial reasoning, \textsc{SpatialVLA}, \textsc{RoboRefer}, Spatial Forcing, and state-free policies incorporate 3D features, depth-aware encoders, and geometry-aware representations~\cite{arxiv2025_qu_spatialvla,arxiv2025_zhou_roborefer,arxiv2025_li_spatialforcing_vla,zhao2025nostate,Li2025BridgeVLAIA}.
However, these methods still rely on natural language as the sole grounding interface, limiting fine-grained spatial reference in cluttered scenes.

\paragraph{Visual prompting methods.}
Visual-prompted pretraining methods such as \textsc{MAGMA}, \textsc{PIVOT}, TrackVLA, TRACE-VLA, \textsc{VIP}, and Embodied One-Vision~\cite{yang2025magma,icml2024_nasiriany_pivot,Wang2025TrackVLAEV,zheng2024tracevla,icml2025_li_vip,arxiv2025_qu_embodiedonevision} use trajectories, sketches, or egocentric observations as auxiliary supervision, but primarily improve representation learning rather than providing explicit spatial reference at inference.
Visual-instruction policies like \textsc{VIMA} and \textsc{Interleave-VLA}~\cite{jiang2023vima,corlws2025_fan_interleavevla} interleave image and text tokens for manipulation, while \textsc{RoboGround}, \textsc{MOKA}, RoVI, and other prompting interfaces use bounding boxes, marks, or geometric primitives as direct visual instructions~\cite{cvpr2025_huang_roboground,rss2024_fang_moka,cvpr2025_li_rovi,case2025_muttaqien_act_visual_prompt}.
Unlike VLM grounding, which outputs text and is evaluated by semantic accuracy, VLA grounding must produce precise motor trajectories for successful manipulation, requiring action-level spatial precision, temporal consistency across manipulation sequences, and plug-and-play compatibility with existing VLA architectures without task-specific modification.
However, existing approaches either lack explicit language-to-region binding or rely on costly annotations.
Our method directly anchors linguistic referents to pixels within a unified VLA policy.

\section{METHOD}
\label{sec:method}

\vspace{-1mm}

In this section, we introduce the core design of our \ourvla, including the construction of visually grounded instructions, the automatic data annotation pipeline, and the co-training strategy with text-only instructions. A pipeline of the inference is illustrated in Figure \ref{fig:method-inference}.

\vspace{-2mm}

\subsection{Preliminary}
A standard VLA policy maps visual observations and textual instructions to an action distribution for embodied control.
Formally, at time step $t$, given visual observations $\mathbf{I}_t = \{I^1_t, I^2_t, \ldots, I^n_t\}$ from multiple cameras and text instruction $l_t$, the policy predicts the next robot action $\hat{a_t}$ as:

\begin{equation}
    \hat{a_t} = \pi_\theta(l_t, \mathbf{I}_t),
\end{equation}
where $\pi_\theta$ denotes the parameterized VLA model.
In this formulation, visual observations provide scene context while object and task references are derived from text instruction.
Once linguistic descriptions become ambiguous or incomplete, the VLA lacks explicit grounding to localize the intended target, leading to referential errors.

\subsection{Visually Grounded Instruction}
In real-world manipulation, language alone is often insufficient for precise referring in several common situations.
These include targets embedded in heavy clutter with many visually similar items, fine-grained spatial goals such as a specific slot or point on a plain surface without visual anchors, and objects whose appearance or category name is rare or unfamiliar to the model.
These cases expose the limitation of purely linguistic instructions for conveying precise spatial and referential intent.

To overcome these limitations, we augment the standard VLA interface with a visually grounded instruction. We define the visual grounding as a spatial marker $g \in \mathcal{G}$ applied to the first-frame image $I^1_0$, where $\mathcal{G}$ represents the space of possible grounding formats (bounding boxes, masks, coordinates, etc.). This yields the grounded image $\tilde{I}_{g,0} = \mathcal{M}(I^1_0, g)$, where $\mathcal{M}$ is the marker overlay function.

The visually grounded instruction is then formulated as a tuple:
\begin{equation}
    \mathcal{I}_{\text{VG}} = (l_t, \tilde{I}_{g,0}),
\end{equation}
where $l_t$ expresses only high-level intent (e.g., \textit{pick up}, \textit{place}), while all target-specific spatial information is encoded in $\tilde{I}_{g,0}$.

We explore multiple grounding formats, including bounding boxes, masks, and textual coordinates. We use a bounding-box overlay as the default visual prompt in our implementation. Other formats remain compatible and can be used zero-shot at inference, and we provide detailed prompt formulations and usage examples in the appendix.

The bounding box on $\tilde{I}_{g,0}$ fully determines the target, without any linguistic cues about its identity or spatial relation.
Although we instantiate visual grounding as bounding boxes in our experiments, the interface is agnostic to the specific marker shape and can also support other lightweight visual cues such as circles or clicks.

Conditioned on this instruction format, the policy at any time step $t$ predicts the next action as
\begin{equation}
    \hat{a_t} = \pi_\theta(l_t, \mathbf{I}_t, \tilde{I}_{g,0}),
\end{equation}
where $\mathbf{I}_t$ denotes the current multi-view observations.
This plug-and-play extension directly associates textual intent with pixel-level evidence, enabling the model to uniquely identify the referred target without modifying the underlying VLA backbone.

\input{figures/task_overview/task_overview}

\subsection{Data Preparation}
\paragraph{Data Annotation Pipeline}
While visually grounded instructions enable precise object referring, collecting bounding boxes for every episode by hand is prohibitively expensive.
To make training scalable, we employ Multi-modal Large Language Models (MLLMs) to automatically generate the grounding signal.
The pipeline of the training workflow is illustrated in Figure \ref{fig:method-training}.

Concretely, for each demonstration we adopt a four-stage automatic annotation pipeline.
First, the MLLM performs video-level scene understanding conditioned on the full episode and its textual description.
Second, it selects one or more key frames where the target object is clearly visible.
Third, the MLLM predicts a bounding box for the referred target on the selected key frame.
Finally, this box is propagated to the first-frame overhead image $\tilde{I}_{g,0}$, yielding the $(\tilde{I}_{g,0}, g)$ pairs used as supervision for the grounded instruction branch described above.
We adopt this single-frame grounding strategy primarily for efficiency: performing visual grounding once per episode avoids redundant per-frame processing and significantly accelerates both data generation and inference.
Empirically, we find that the generated boxes provide sufficiently accurate supervision for diverse manipulation tasks.
The full prompt, structured JSON output, and annotation quality analysis are included in Appendix~\ref{sec:annotation_pipeline}.

\paragraph{Data Augmentation}
To improve robustness of the visual grounding signal, we apply two simple augmentations directly to the grounded-image input:
\begin{itemize}
    \item \textbf{Random translation.} We randomly translate the grounded image so that the scene and box move together, encouraging the policy to rely on the target's \emph{relative} position in the scene rather than absolute pixel coordinates.
    \item \textbf{Localized CutMix~\cite{yun2019cutmix}.} Within the bounding box, we partially replace the object appearance with ImageNet~\cite{deng2009imagenet} patches while leaving the surrounding context unchanged, preventing the model from overfitting the grounding prompt to a small set of seen object instances.
\end{itemize}
Formally, the augmented grounded image is obtained through:
\begin{equation}
    \tilde{I}'_{g,0} = \mathcal{A}_{\text{trans}}(\mathcal{A}_{\text{mix}}(\tilde{I}_{g,0}, g), \delta),
\end{equation}
where $\mathcal{A}_{\text{mix}}$ applies localized CutMix within the bounding box region, $\mathcal{A}_{\text{trans}}$ applies random translation with offset $\delta$, and the composition ensures both spatial and appearance invariance.
These augmentations promote generalization under spatial perturbations and novel object appearances without changing the underlying policy architecture.

\subsection{Training and Inference}
\label{subsec:training}

\paragraph{Co-training with Text-only Instructions}
To maintain compatibility with conventional text-based policies, we co-train \ourvla\ on a balanced mixture of pure text instructions and text-plus-visual-grounding instructions with a 1:1 ratio.
For text-only samples, the policy reduces to the standard VLA setting and conditions only on $(l_t, \mathbf{I}_t)$, while for visually grounded samples the model additionally receives the first-frame grounded image $\tilde{I}_{g,0}$ and its visual marker $g$.

We formally define the two instruction modalities as collections of paired training samples:
\begin{align}
\mathcal{D}_{\text{text}} &= \{ (l_t, \mathbf{I}_t) \}, \qquad
\mathcal{D}_{\text{visual}} = \{ (l_t, \mathbf{I}_t, \tilde{I}_{g,0}, g) \}.
\label{eq:two_sets}
\end{align}

The full co-training dataset is the union of these two modalities:
\begin{equation}
\mathcal{D}
= \mathcal{D}_{\text{text}} \,\cup\, \mathcal{D}_{\text{visual}}.
\label{eq:full_dataset}
\end{equation}

As shown in Equ~\eqref{eq:full_dataset}, this unified corpus allows the model to benefit from both modalities: visually grounded samples improve spatial disambiguation, while text-only samples preserve strong instruction-following behavior when explicit visual grounding is unnecessary.

\paragraph{Training Configuration}
We fine-tune the $\pi_{0.5}$ backbone for 20k steps per task with batch size 512.
We use AdamW optimizer with learning rate $5 \times 10^{-5}$ and constant schedule.
The training objective is the flow-matching loss from the base model.
The text-to-visual-grounding ratio in our training data is 1:0.42, ensuring the model maintains strong text-only capabilities while learning visual grounding.

\paragraph{Interactive Visual Grounding at Inference Time}
At inference time, \ourvla supports two complementary forms of interactive visual grounding.
The first mode offers highly interactive and precise control: the user directly manipulates a GUI that displays the overhead view, draws a bounding box on the desired target, and optionally provides a short textual command (e.g., "pick up" or "place here"), producing a clear and unambiguous grounded instruction $(l_t, \tilde{I}_{g,0})$.
The second mode enables natural, human-like interaction: a MLLM observes the scene together with human pointing or other gestural cues and automatically predicts a bounding box around the indicated object, which is then combined with the user's verbal command to form the grounded instruction.
Both modes realize intuitive point-to-act control while reusing the same VG-VLA policy trained with visually grounded instructions.
Appendix~\ref{sec:inference_modes} details both interaction modes and the associated MLLM prompt.

\section{EXPERIMENTS}

We comprehensively evaluate \ourvla\ across a suite of real-world robotic manipulation tasks designed to probe its ability to resolve visual–linguistic ambiguity. This section introduces our experimental setup, evaluation results, and ablation studies.

\subsection{Setup}
\label{sec:exp_setup}
\paragraph{Real-world Task}
We evaluate on six manipulation tasks with challenging referring instructions (Figure~\ref{fig:task_overview}c):
\begin{itemize}
    \item \textbf{Target object referring.}
    \begin{itemize}
        \item \textit{Irregular-shape picking:} ``Pick the clay piece with red and blue stripes, round on top and square on bottom, on the left side of the pile'' from cluttered irregular objects.
        \item \textit{OOD object picking:} ``Pick the battery'' or ``pick the remote control'' from eight unseen items.
    \end{itemize}
    \item \textbf{Target location referring.}
    \begin{itemize}
        \item \textit{Cluttered picking:} ``Pick the bottle to the right of the leftmost bottles, in the middle of the desk'' among eight identical bottles.
        \item \textit{Egg-slot picking:} ``Pick the egg in row 2, column 3'' from a 4$\times$4 tray.
        \item \textit{Plain placement:} ``Place the cup 15cm to the right and 10cm forward'' on a featureless tabletop.
        \item \textit{Egg-slot placement:} ``Place into row 3, column 1'' of an egg tray.
    \end{itemize}
\end{itemize}
Each task includes 1-2 scene configurations, totaling 12 evaluation scenes with 30 trials per scene. Detailed text-only prompts for all test scenarios are provided in Appendix~\ref{sec:test_scenarios}.

\input{tables/main_results}

\paragraph{Evaluation Metric.}
Each task is evaluated over 30 trials. Success requires grasping and lifting the target (picking) or positioning within the goal region (placing). Each trial permits two retries. Failure is recorded when all attempts fail, timeout exceeds 30s, or placement error exceeds 10cm.

\begin{figure}[t]
    \centering
    \includegraphics[width=0.8\columnwidth]{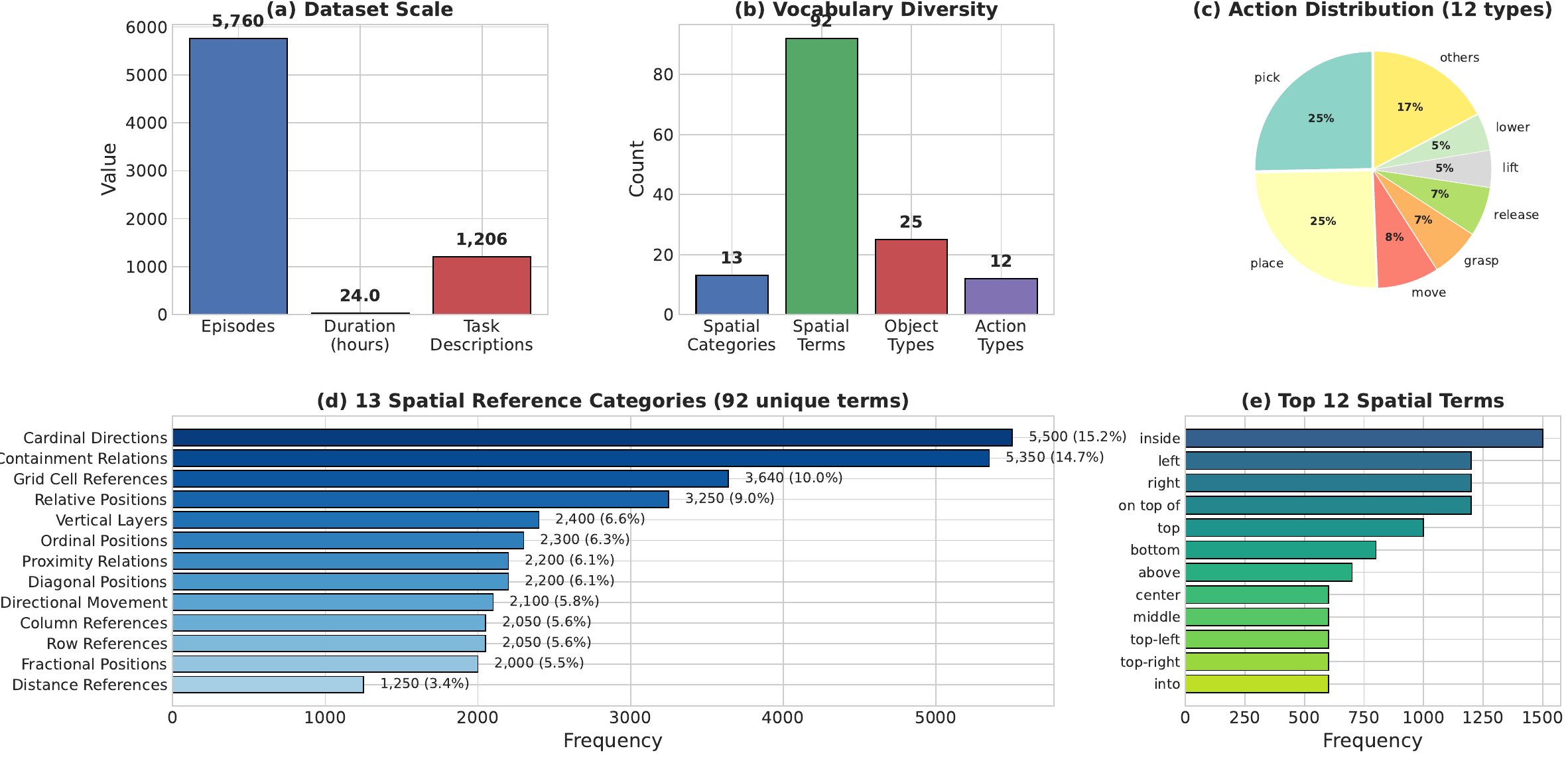}
    \vspace{2pt}

    \includegraphics[width=0.8\columnwidth]{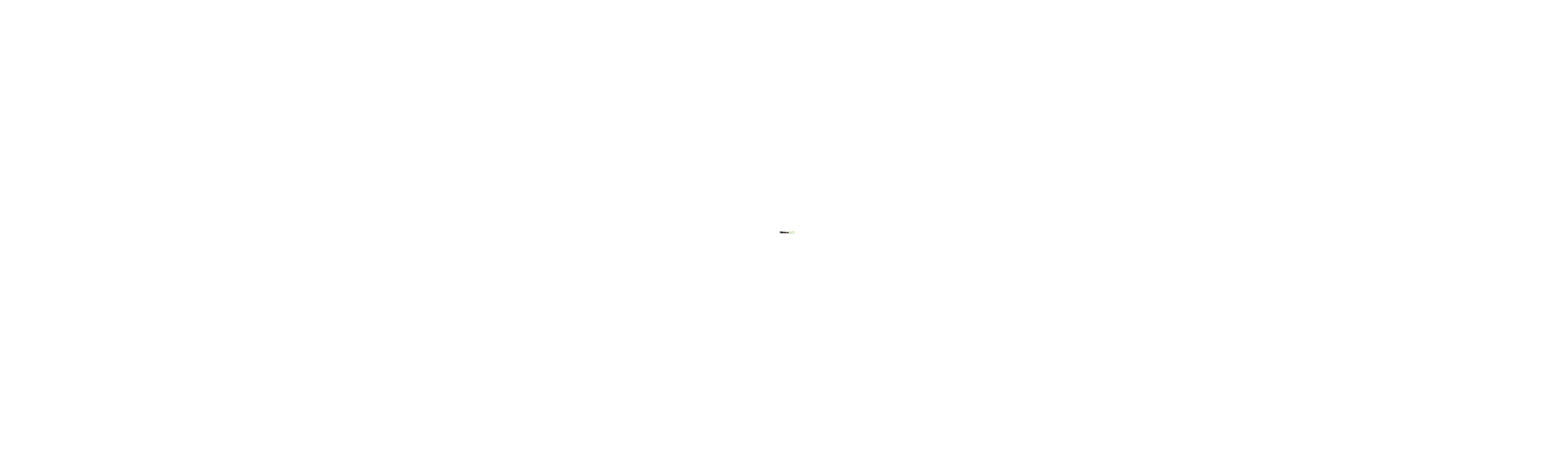}
    \vspace{-3pt}
    \caption{Dataset statistics. (Top) Object and action category distributions showing diverse manipulation scenarios. (Bottom) Spatial reference word cloud illustrating the variety of linguistic spatial expressions in our training data.}
    \label{fig:dataset_stats}
    \vspace{-5mm}
\end{figure}

\paragraph{Real-world Data.}
We collect $\sim$2 hours of demonstrations per scene across 12 scenes. All textual annotations are manually verified, guaranteeing fair comparison across different instruction modalities. Figure~\ref{fig:dataset_stats} shows object/action distributions and spatial reference word frequencies, demonstrating data diversity.

\paragraph{VLA Model Backbone and Robot Embodiment}
We adopt $\pi_{0.5}$ as the backbone, fine-tuning for 20k steps per task with identical optimization settings. Evaluations use a dual-arm robot with one overhead and two wrist cameras. To demonstrate plug-and-play capability, we also evaluate on $\pi_{0}$ and a full-body humanoid robot (Figure~\ref{fig:task_overview}).

\begin{figure}[t]
    \centering
    \includegraphics[width=0.9\columnwidth]{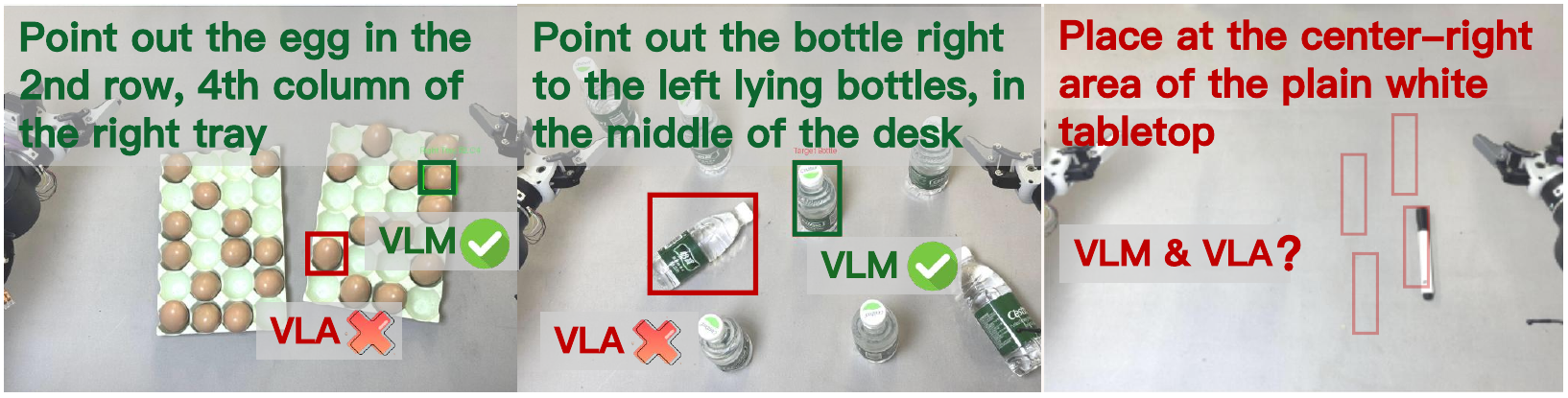}
    \caption{VLM localization vs VLA execution gap. \textbf{Left two examples}: VLMs successfully localize targets with detailed text descriptions (60-70\% accuracy), but text-only VLAs fail to execute (25\% success), revealing a language-action alignment gap. \textbf{Right example}: Language fundamentally cannot describe the target (plain surface without anchors), and even the strongest VLMs fail to localize. Point-VLA addresses both scenarios through visual grounding.}
    \vspace{-3mm}
    \label{fig:vlm2vla_gap}
\end{figure}

\paragraph{On Information Equivalence and Comparison Fairness}
This work addresses language's fundamental limitations in spatial referring. Unlike prior work on simple lab scenarios, we evaluate on \emph{extreme real-world challenges} representing language's boundary cases in daily life. To ensure fairness, text instructions are designed as \emph{maximum information with minimum description}. Our design distinguishes two scenarios. \textbf{(1) Language-describable tasks}: Both modalities carry \emph{identical information} in different formats. VLM benchmarks (Gemini-2.5, Qwen-VL-Max) achieve $\sim$70\% localization for cluttered picking and $\sim$60\% for matrix layouts (Figure~\ref{fig:vlm2vla_gap}), proving language is unambiguous. Yet text-only VLAs achieve only 25\% success while Point-VLA achieves 95\%. Critically, our text baseline is trained on fine-grained positional references with rich directional expressions (Figure~\ref{fig:dataset_stats}), making this the fairest comparison. The gap reveals VLA's language-action alignment bottleneck at language's boundary, not information insufficiency. \textbf{(2) Language-inexpressible tasks}: Even VLMs cannot localize on plain surfaces. Converting visual grounding to text coordinates fails (Table~\ref{tab:visual_form_ablation}). By comparing grounded modes at language's edge cases, we highlight our method's innovation in solving previously intractable real-world challenges.

\subsection{Main Evaluation}
We evaluate \ourvla on six real-world robot manipulation tasks introduced in Section~\ref{sec:exp_setup}, comparing it against two baselines:
(1) a \textbf{text-instruction VLA} that relies solely on linguistic instructions (no explicit visual grounding), and
(2) \textbf{Interleave}~\cite{corlws2025_fan_interleavevla}, which mixes image patches with text tokens but lacks explicit spatial grounding and therefore undermines precise positional understanding.

As shown in Table~\ref{tab:main_evaluation}, \ourvla\ achieves the highest success rate across all tasks.
It yields average absolute gains of \textbf{+60.1} and \textbf{+52.5} percentage points over the text-only and Interleave baselines, respectively.
These results establish \ourvla\ as substantially more robust in both object-level and spatial understanding.

Although Interleave can slightly aid unseen-object reference through coarse visual appearance cues, it lacks positional encoding and struggles with directional or layout-sensitive instructions.
Consequently, it often performs comparably or worse than the text-only baseline on tasks requiring precise spatial understanding.

In contrast, \ourvla\ learns explicit \emph{positional reference binding} through bounding-box grounding, enabling accurate and transferable spatial understanding.
Since our training corpus already covers a broad set of common directional expressions (e.g., ``left,'' ``top-right,'' ``next to''), the comparison remains fair:
the advantage arises not from additional linguistic exposure but from the model's ability to generalize spatial understanding to cluttered and visually complex environments.

This demonstrates that pixel-level visual grounding in \ourvla\ provides a more robust and generalizable learning signal,
allowing the model to resolve challenging scenarios with ambiguous or linguistically indescribable spatial configurations that text-only or token-interleaved methods cannot handle.

\subsection{Compatibility with Text Instructions}
\input{figures/experiment/text_compatible}
We co-train \ourvla\ on a balanced mixture of text-only and visually grounded instructions, as described in Section~\ref{subsec:training}.
To evaluate the effect of this co-training, we compare three instruction modes on three location-referring scenarios: the Text VLA baseline, \ourvla evaluated with text-only instructions (\ourvla($l$)), and \ourvla evaluated with visually grounded instructions (\ourvla(VGI), where language gives only the high-level action and the bounding box provides all spatial reference).

\begin{itemize}
    \item \textbf{Relative position referring:} the target is specified by relative direction, e.g., ``pick the object in the top-left corner.''
    \item \textbf{Matrix layout referring:} multiple identical objects are arranged in rows and columns, and the instruction specifies the target by its index, e.g., ``pick the object in the second row, third column.''
    \item \textbf{Reference-based:} the target is located relative to another object, e.g., ``pick the cup to the right of the bowl.''
\end{itemize}

As shown in Figure~\ref{fig:text_compatible}, \ourvla($l$)—the same Point-VLA policy evaluated \emph{only} with text instructions—achieves higher success rates than the Text VLA across all three scenarios.
This indicates that co-training with visually grounded data improves the underlying VLA policy itself, so that even when used in a text-only mode it follows linguistic spatial references more accurately than a purely text-trained baseline.
Importantly, these improvements demonstrate that visual grounding training provides genuine capability enhancement rather than task simplification. The model learns better spatial reasoning and language-action alignment that transfers to text-only scenarios, proving that \ourvla is not merely ``following the red box'' but developing deeper understanding of spatial references and manipulation semantics.

\subsection{Plug-and-Play Point-VLA}
\input{tables/plug-and-play}
To assess plug-and-play generalization, we evaluate \ourvla\ across policy backbones and embodiments using the $\pi_{0.5}$ and $\pi_{0}$ models, together with a full-body humanoid robot with active waist and leg control.
The task suite includes tabletop manipulations (cluttered picking, container placing) and whole-body desk-organization activities (retrieving objects such as pens, remotes, or docking hubs and placing them into designated containers), as summarized in Table~\ref{tab:plug_and_play}.

Averaged over cluttered picking and container placing, \ourvla\ consistently delivers substantial improvements over the corresponding $l$-instruction baselines across all three settings in Table~\ref{tab:plug_and_play}, demonstrating reliable gains across both the original $\pi_{0.5}$ backbone and the lighter $\pi_{0}$ model.
Under substantial embodiment and viewpoint changes, \ourvla\ also maintains strong spatial grounding and reliable target localization on the full-body humanoid, demonstrating that visually grounded instructions transfer effectively across architectures and embodiments.
Taken together, these results show that \ourvla\ functions as a modular, architecture-agnostic policy interface that generalizes across embodiments, task families, and larger-scale scenes, supporting reliable deployment of vision–language–action models in everyday manipulation.

\input{figures/visual_form/visual_form}

\input{tables/VG_Form}

\subsection{Ablation on the Visually Grounded Instruction Formulation}

To examine how different grounding signals affect instruction understanding,
we compare three formulations of visually grounded input, as shown in Figure~\ref{fig:visual_form}:
\begin{itemize}
    \item \textbf{Textual coordinates.} The numerical coordinates of the bounding box are appended to the instruction text (e.g., ``$x_1=40, y_1=80, x_2=150, y_2=220$'').
    \item \textbf{Object-only masking.} Only the target object region is retained, while the rest of the image is masked out.
    \item \textbf{Bounding-box overlay (ours).} The target region is explicitly highlighted on the input image using a visual bounding box overlay.
\end{itemize}

We evaluate these variants on four representative real-world tasks—regular and disturbed egg-tray picking, cluttered tabletop picking, and plain placement—summarized in Table~\ref{tab:visual_form_ablation}.

Qualitative inspection shows textual-coordinate formulations overfit absolute positions, failing when the tray shifts or camera view changes; mask-based variants remove contextual information, causing confusion with similar objects.
The bounding-box overlay preserves global context while providing explicit local reference.
Critically, Box-text contains \emph{identical spatial information} to box overlay yet performs significantly worse (e.g., 37\% vs 80\% on disturbed egg picking, 73\% vs 95\% on plain placement), confirming Point-VLA's advantage stems from visual representation format rather than information quantity.
These results indicate effective visual grounding requires balancing \textit{local precision} with \textit{global contextual awareness}, which bounding-box overlays achieve.

\vspace{-2mm}

\subsection{Ablation on Data Augmentation}

\vspace{-2.2mm}

\input{tables/data_aug}

We ablate random translation and CutMix by removing each during fine-tuning (Table~\ref{tab:ablation_augmentation}). Removing random translation substantially hurts egg-task performance (where the tray shifts up to 10\,cm at execution) but leaves OOD results unchanged, indicating \ourvla\ learns object location relative to the environment rather than absolute image coordinates. Removing CutMix markedly hurts OOD performance but leaves egg-task results similar, indicating mixed visual contexts improve robustness to novel appearances; without it, the model overfits training appearances rather than relying on spatial grounding. Both augmentations strengthen spatial understanding and visual grounding priors without extra data collection.

\paragraph{Object-Level vs Position-Level Grounding}
Perturbation experiments verify \ourvla\ learns object-level grounding rather than memorizing pixel coordinates: when the egg tray shifts up to 10cm after annotation, \ourvla\ maintains 80\% success, demonstrating grounding on visual features and relative spatial context. This explains partial grasping ability (e.g., cup handles): the bounding box specifies ``what to grasp'' while ``how to grasp'' is learned from demonstrations.

\subsection{Scaling Behavior with Training Data}

\input{figures/experiment/scaling}

We study how performance scales with training data on OOD object-picking success across ten unseen objects and multiple spatial setups, including tabletops and shelves with varied shapes and colors, as summarized in Figure~\ref{fig:scaling}.
While the Text VLA shows limited gains as data scale grows, \ourvla\ continues to improve following a smoother scaling trend in the log-data regime.
These results indicate that explicit visual grounding provides stronger compositional generalization and higher data efficiency: rather than saturating early like the text-only baseline, \ourvla\ continues to benefit from increased training data, confirming that pixel-level grounding amplifies the scaling behavior of VLAs while maintaining robust OOD generalization.

\section{CONCLUSION AND LIMITATIONS}

Point-VLA demonstrates that explicit visual grounding is a practical plug-and-play interface for VLAs, resolving referential ambiguity beyond text-only instructions while improving robustness, generalization, and text-only performance with scalable supervision. Its current formulation, however, uses first-frame-only grounding for efficiency and mainly box-based prompts, which can limit performance under large viewpoint changes, target motion, long-horizon interaction, and visually complex or heavily occluded scenes.

\vspace{-1mm}

\bibliographystyle{IEEEtran}
\bibliography{references}


\clearpage
\appendix
\input{appendix}

\end{document}

%% file: figures/method/inference.tex
\begin{figure*}[t]
    \centering
    \includegraphics[width=\textwidth]{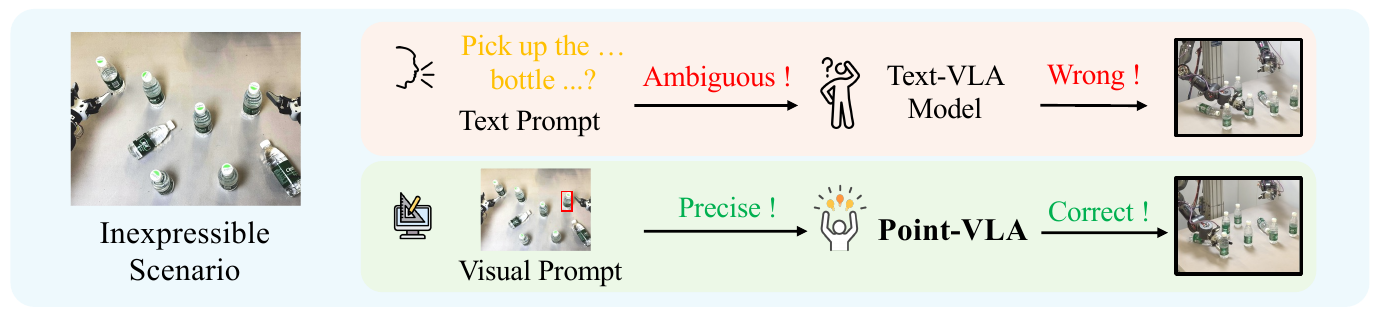}
    \caption{Point-VLA resolves linguistically inexpressible references through explicit visual grounding. In scenes with many visually similar objects, even complex and fully specified textual descriptions cannot generalize reliably, causing ambiguous and incorrect actions.}
    \vspace{-6mm}
    \label{fig:method-inference}
\end{figure*}

%% file: figures/method/training.tex
\begin{figure*}[t]
    \centering
    \includegraphics[width=0.8\textwidth]{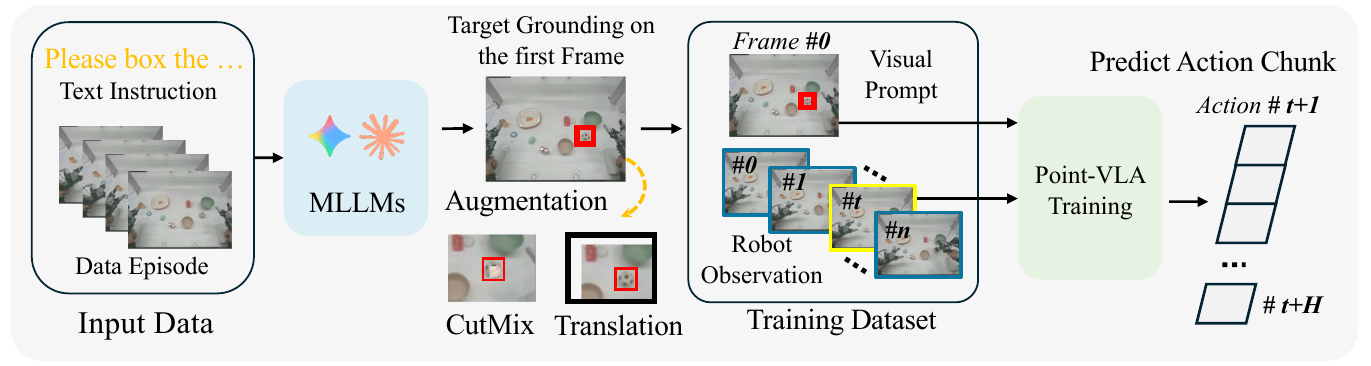}
    \caption{We obtain the visual prompt by drawing a bounding box on the first frame, either annotated automatically or manually. This grounded frame is then lightly augmented (CutMix, translation) and paired with every robot observation in the episode. The model is trained using both the current observation and the fixed grounded first-frame prompt, enabling consistent pixel-level target grounding throughout the trajectory.}
    \vspace{-5mm}
    \label{fig:method-training}
\end{figure*}

%% file: figures/task_overview/task_overview.tex
\begin{figure*}[t]
    \centering
    \includegraphics[width=0.8\textwidth]{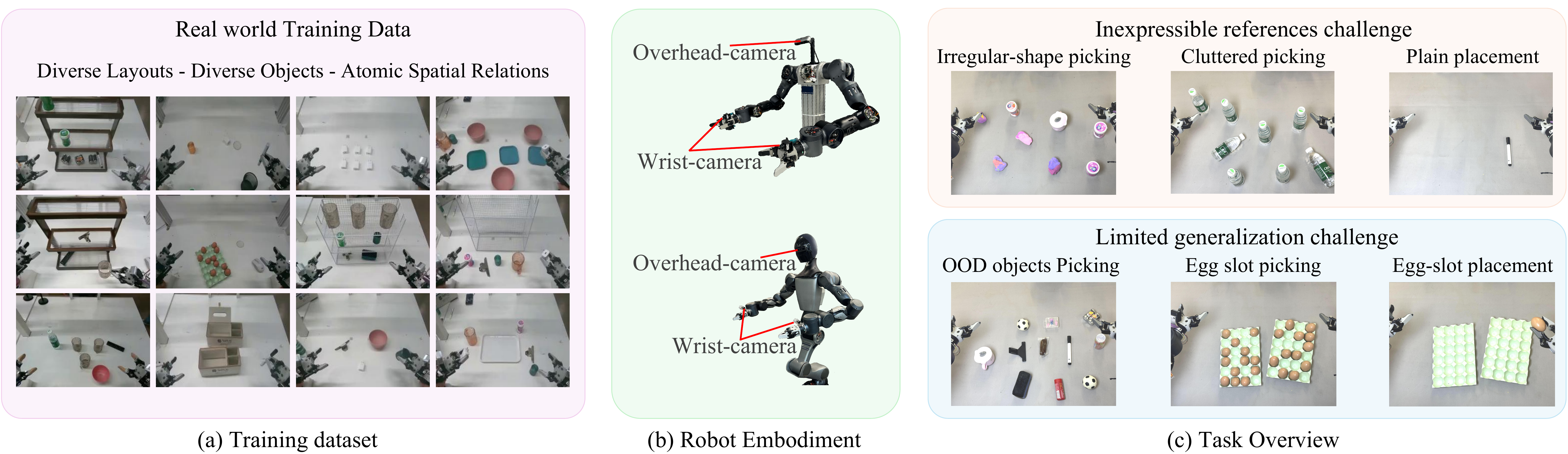}
    \caption{Overviews of our task and robot embodiment. (a) We hire professional operators to collect real-world robot demonstration data. (b) Two robot embodiments used for evaluation: a fixed dual-arm robot and a full-body humanoid robot. (c) Representative tasks, including picking irregular objects, picking OOD objects, picking in clutter, precise picking in dense trays, placing on a plain tabletop without reference points, and precise placing.
    These tasks contain targets that cannot be precisely referred to using text alone.}
    \vspace{-5mm}
    \label{fig:task_overview}
\end{figure*}

%% file: tables/main_results.tex
\begin{table*}[t] \small
\centering
\setlength{\tabcolsep}{5pt}
\renewcommand{\arraystretch}{1.1}
\caption{Success rates (\%) on six real-world manipulation tasks. \textbf{Input modalities}: Text VLA uses detailed linguistic descriptions (e.g., "pick the egg in row 2, column 3"); Interleave-VLA uses text + ungrounded image patches; Point-VLA uses minimal text (e.g., "pick") + visual bounding box. For language-describable tasks (Irregular, OOD, Clutter, Egg pick), Text VLA and Point-VLA receive equivalent information in different formats. For language-inexpressible tasks (Plain placement), only visual grounding can specify the target.}
\label{tab:main_evaluation}
\begin{tabular}{lccccccc}
\toprule
\multirow{2}{*}{Method} & \multicolumn{4}{c}{Pick} & \multicolumn{2}{c}{Place} & \multirow{2}{*}{Avg} \\
\cmidrule(lr){2-5} \cmidrule(lr){6-7}
& Irregular object & OOD object & Clutter scenario & Egg from slot & Plain tabletop & Egg into slot & \\
\midrule
Text VLA & 30.0 & 57.5 & 43.3 & 10.0 & 30.0 & 23.3 & 32.4 \\
Interleave-VLA & 60.0 & 86.7 & 33.3 & 13.3 & 26.7 & 20.0 & 40.0 \\
\ourvla (ours) & \textbf{96.7} & \textbf{92.5} & \textbf{94.3} & \textbf{86.7} & \textbf{95.0} & \textbf{90.0} & \textbf{92.5} \\
\bottomrule
\end{tabular}
\vspace{-10pt}
\end{table*}

%% file: figures/experiment/text_compatible.tex
\begin{figure}[htbp]
\centering
\vspace{-2mm}
\includegraphics[width=0.8\linewidth]{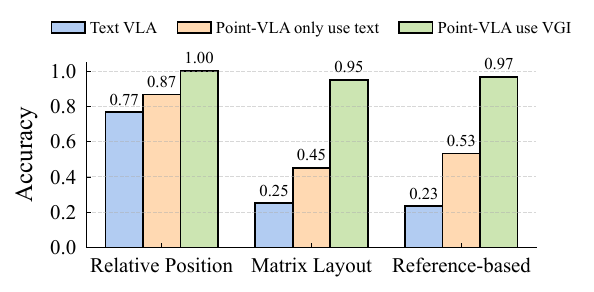}
\vspace{-3mm}
\caption{Success rates (\%) on three spatial referring tasks under three instruction modes: the Text VLA baseline, \ourvla($l$) with text-only instructions, and \ourvla(VGI) with visually grounded instructions (text for high-level action, bounding box for spatial reference).
\ourvla($l$) matches or exceeds the baseline, and \ourvla(VGI) achieves the highest success rates on complex spatial references.}
\label{fig:text_compatible}
\vspace{-2mm}
\end{figure}

%% file: tables/plug-and-play.tex
\begin{table}[t]
\centering
\small
\setlength{\tabcolsep}{2.5pt}
\renewcommand{\arraystretch}{1.1}
\caption{Success rates (\%) across different VLA backbones and robot embodiments. \ourvla consistently outperforms the corresponding text-instruction VLAs under different model and embodiment settings, demonstrating that \ourvla transfers reliably across architectures and robots after lightweight fine-tuning.}
\label{tab:plug_and_play}
\begin{tabular}{llccc}
\toprule
\multirow{2}{*}{Backbone} & \multirow{2}{*}{Robot}      & \multirow{2}{*}{Method} & Clutter & Container \\
                       &                            &                        & Picking & Placing \\ \midrule
\multirow{2}{*}{$\pi_{0.5}$} & \multirow{2}{*}{dual-arm}  & Text VLA & 43.3 & 30.0 \\
                       &                            & Point-VLA & 94.3 & 93.3 \\ \midrule
\multirow{2}{*}{$\pi_{0.5}$}   & \multirow{2}{*}{full-body} & Text VLA & 41.7 & 33.3 \\
                       &                            & Point-VLA & 83.3 & 76.6 \\ \midrule
\multirow{2}{*}{$\pi_{0}$} & \multirow{2}{*}{dual-arm}  & Text VLA & 20.0 & 13.3 \\
                       &                            & Point-VLA & 63.3 & 56.6 \\ \bottomrule
\end{tabular}
\vspace{-3mm}
\end{table}

%% file: figures/visual_form/visual_form.tex
\begin{figure}[htbp]
    \centering
    \includegraphics[width=0.85\linewidth]{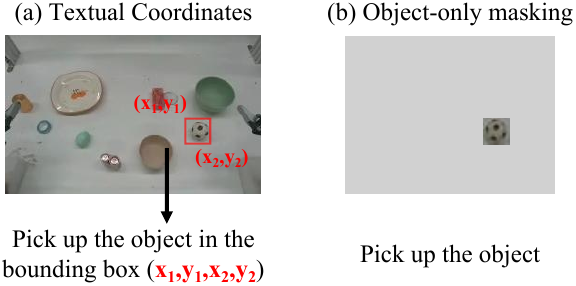}
    \caption{Illustration of two alternative forms of visually grounded instructions. (a) Bounding-box coordinates inserted into the text instruction; (b) Only the region inside the bounding box is kept, with the remaining image masked out.}
    \label{fig:visual_form}
\end{figure}

%% file: tables/VG_Form.tex
\begin{table}[t]
\centering
\small
\setlength{\tabcolsep}{3.5pt}
\renewcommand{\arraystretch}{1.05}
\caption{Ablation on the form of visually grounded instructions.
We compare a text-only baseline and three visually grounded representations: textual coordinates (\textit{box-text}), object-only masking, and image with bounding-box overlay (used in \ourvla). The bounding-box overlay achieves the best results across all tasks, while the other two forms provide only limited improvement and may even underperform text-only instructions in certain cases.}
\label{tab:visual_form_ablation}
\begin{tabular}{lcccc}
\toprule
\textbf{Method} &
\makecell{{Egg pick}} &
\makecell{{Egg pick}$^{\dagger}$} &
\makecell{{Cluttered}\\pick} &
\makecell{{Plain}\\placement} \\
\midrule
Text VLA & 10 & 13 & 43.3 & 30 \\
Box-text & 70 & 37 & 83 & 73 \\
Mask & 43 & 10 & 73 & 77 \\
Box overlay (ours) & \textbf{86.7} & \textbf{80} & \textbf{94.3} & \textbf{95.0} \\
\bottomrule
\end{tabular}
\vspace{2pt}
{\scriptsize $^{\dagger}$ Disturbed tray: egg-tray randomly shifted during execution.}
\vspace{-10pt}
\end{table}

%% file: tables/data_aug.tex
\begin{table}[htbp]
\centering
\small
\setlength{\tabcolsep}{2.5pt}
\renewcommand{\arraystretch}{1.1}
\caption{Ablation on data augmentation. We ablate box random translation and CutMix by removing them individually during training.}
\label{tab:ablation_augmentation}
\begin{tabular}{lcccc}
\toprule
 & Egg pick$^{\dagger}$ & Egg place$^{\dagger}$ & OOD pick & OOD place \\
\midrule
\textbf{Point-VLA} & \textbf{80.0} & \textbf{76.6} & \textbf{92.5} & \textbf{90.0} \\
w/o shift & 20.0 & 15.0 & 95.0 & 90.0 \\
w/o cutmix & 80.0 & 73.3 & 60.0 & 45.0 \\
\bottomrule
\end{tabular}
\vspace{2pt}
{\scriptsize \\$^{\dagger}$ Evaluated with the egg tray randomly shifted by up to 10\,cm during execution.}
\end{table}

%% file: figures/experiment/scaling.tex
\begin{figure}[t]
\centering
\includegraphics[width=0.7\linewidth]{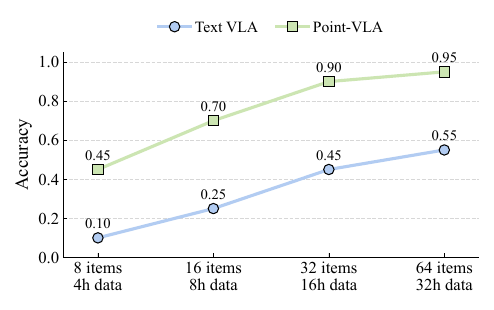}
\caption{Scaling with object diversity. Accuracy improves as training object diversity increases, showing that \ourvla continues to benefit from broader visual variation.}
\vspace{-5mm}
\label{fig:scaling}
\end{figure}

%% file: appendix.tex
\subsection{Automatic Annotation Pipeline Details}
\label{sec:annotation_pipeline}

To construct visually grounded instructions at scale, we use the Gemini ER1.5 multimodal LLM with a single carefully engineered prompt that operates directly on multi-view frame sequences.
For each demonstration episode, we uniformly sample $T=20$ frames from three synchronized cameras: an overhead high-view camera and two wrist cameras mounted on the left and right arms, then package them together with the natural-language task description $l_t$ into one multimodal query.
The prompt asks the MLLM to (1) classify the manipulation as a pick or place task and determine which arm is active, (2) identify the key gripper-action moment in the corresponding wrist-camera stream, (3) localize the manipulated object or receiving container in the overhead view with a tight bounding box, and (4) summarize these decisions with explicit step-by-step reasoning in a constrained JSON format.
These reasoning operations are organized into four conceptual stages of our automatic annotation pipeline, as detailed below.

\paragraph{Stage 1: Multi-view episode understanding.}
The MLLM jointly analyzes all three views and the task text to decide whether the episode is a pick or place task and which arm (left or right) is performing the manipulation.
This coarse understanding constrains the subsequent search to the corresponding wrist camera and to the appropriate target type (object versus container).

\paragraph{Stage 2: Key moment localization.}
Conditioned on the chosen arm, the model focuses on the associated wrist-view frames and identifies the moment when the gripper closes to pick an object or opens to place it.
This yields a key-frame index in the wrist stream that anchors the temporal reasoning.

\paragraph{Stage 3: Overhead target localization.}
Using the fact that the overhead camera is static, the model then localizes the manipulated object (pick) or receiving container (place) in the high-view frames, leveraging motion cues around the key moment as well as context from earlier and later frames.
The output is a tight 2D bounding box in normalized coordinates.

\paragraph{Stage 4: Structured JSON output.}
Finally, the MLLM is required to summarize its decision process as a chain-of-thought style explanation and to emit the task type, active arm, key-frame index, and bounding box in a strict JSON schema.
We directly parse this JSON as the visually grounded label for each training episode.

\subsection*{MLLM Prompt and Example Output}

We show below the prompt used for Gemini ER1.5 on our pick-and-place episodes.
The same template is reused across scenarios, with only the task description and view counts substituted.

\begin{figure*}[t]
\begin{smallprompt}
Task description: {task}

I am showing you a robotic manipulation task from three different camera views:
High-view camera (0-{num_high-1}), Left wrist camera ({num_high}-{num_high+num_left-1}), and Right wrist camera.

Step 1: Determine Task Type (Pick vs Place) and Active Arm (Left vs Right).

Step 2: Identify the Key Moment.
For PICK: Find when gripper CLOSES. For PLACE: Find when gripper OPENS.
Determine the frame index in the sequence (0-based indexing).

Step 3: Locate the Target Object/Container in the Overhead View.
IMPORTANT: Use MULTIPLE frames to verify the object.
For PICK: Identify the specific object being grasped.
For PLACE: Identify the specific container/receptacle where the object is being placed.

Important details for bounding box:
Coordinates 0-1000 [ymin, xmin, ymax, xmax]. Box should tightly fit target.

Output format (JSON):
{
  "task_type": "pick" or "place",
  "arm_used": "left" or "right",
  "reasoning_step1": "...",
  "key_frame_index": <frame index>,
  "reasoning_step2": "...",
  "bounding_boxes": [ { "box_2d": [ymin, xmin, ymax, xmax], "label": "..." } ],
  "reasoning_step3": "...",
  "container_verification": "For PLACE tasks ONLY: answer explicit verification questions..."
}
Think step by step and follow the JSON schema exactly.
\end{smallprompt}
\caption{Full multiview annotation prompt.}
\label{fig:prompt_full}
\end{figure*}

Below we include a shortened parsed JSON output for a real pick episode:

\begin{figure*}[t]
\begin{prompt}
{
  "episode_id": "002002",
  "segment_id": 0,
  "task": "pick",
  "task_type": "pick",
  "arm_used": "left",
  "key_frame_index": 19,
  "bbox_results": [
    {
      "bbox": [0.411, 0.618, 0.457, 0.732],
      "label": "red object",
      "original_box": [618, 411, 732, 457]
    }
  ],
  "reasoning_step1": "High-view and wrist-camera videos show the left arm approaching and grasping a red object, confirming this is a left-arm pick action.",
  "reasoning_step2": "Frame 30 is the moment where the gripper fully closes on the red object, so it is selected as the key interaction frame.",
  "reasoning_step3": "The grasp location is tracked backward from the contact point. The red object touched by the gripper lies on the left side of the table, below a white mug. Tracking this position through earlier frames leads to a consistent location in frame 0, where the target object is clearly visible."
}
\end{prompt}
\caption{Example parsed output from the automatic annotation pipeline.}
\label{fig:prompt_example}
\end{figure*}

\subsection*{Annotation Quality Analysis}

To estimate the reliability of automatic annotation, we randomly sample 50 demonstration episodes from the training set.
A human annotator evaluates whether the predicted bounding box (a) refers to the correct target and (b) sufficiently covers the visually grounded region.
Under the fixed-overhead setup, the accuracy reaches \textbf{92\% (46/50)}.
Most errors arise from severe occlusion, ambiguous container views, or extremely similar object appearances.

\subsection{Inference Visual Grounding Modes}
\label{sec:inference_modes}

Point-VLA supports two complementary grounding interfaces during inference: (1) an automatic point-to-box pipeline powered by an MLLM, and (2) a direct user interface where users manually draw bounding boxes.
Both modes produce visually grounded instructions $(l_t, \tilde{I}_{g,0})$ that are compatible with the unified Point-VLA policy.

\subsection{Mode 1: Point-to-Box Inference Pipeline}

A human points toward the desired target in the scene.
The system captures the image, including the pointing gesture, and queries the MLLM to infer the corresponding target region.
The MLLM outputs a bounding box that tightly localizes the indicated object.
This box is then paired with a minimal textual command (e.g., ``pick up'') to form the visually grounded instruction used by Point-VLA.

\begin{figure}[t]
    \centering
    \includegraphics[width=\linewidth]{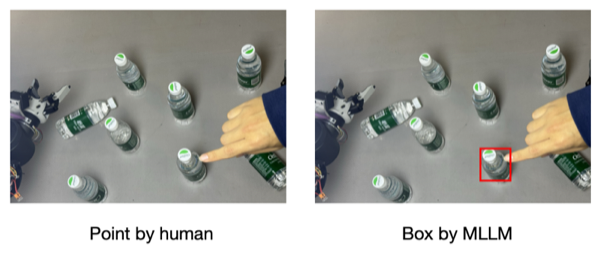}
    \caption{\textbf{Inference Mode 1: Point-to-Box.} A user points at the desired target, and the MLLM converts the gesture into a grounded bounding box that can be directly consumed by Point-VLA.}
    \label{fig:point_to_box}
\end{figure}

\begin{figure}[t]
\begin{prompt}
Please provide the bounding box coordinate of the region this sentence describes: {task}
The format should be as follows: [{"box_2d": [ymin, xmin, ymax, xmax], "label": <label for the object>}] normalized to 0-1000. The values in box_2d must only be integers.

Important:
- Identify the object that the hand is pointing to in the image
- Provide the bounding box coordinates in [ymin, xmin, ymax, xmax] format
- All coordinates should be integers normalized to 0-1000 range
- Include a descriptive label for the detected object
- Output must be valid JSON format
\end{prompt}
\caption{MLLM prompt for Point-to-Box inference.}
\label{fig:prompt_point_to_box}
\end{figure}

This enables a fully automatic point-to-act control interface, requiring no manual annotation at inference time.

\subsection{Mode 2: User Interface for Manual Box Drawing}

Users may instead interact with a simple on-screen UI that shows the current overhead image.
Through mouse dragging, the user draws a bounding box around the desired target and optionally enters a short action phrase (e.g., ``pick up'' or ``place here'').
The resulting pair $(l_t, \tilde{I}_{g,0})$ is passed directly to the Point-VLA policy.
This interface provides pixel-level precision and is particularly useful for data collection, teleoperation, and evaluation in human-in-the-loop settings.

\begin{figure}[t]
    \centering
    \includegraphics[width=\linewidth]{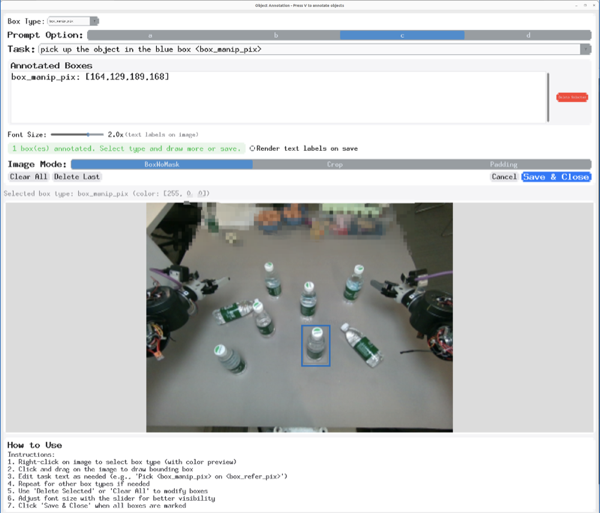}
    \caption{\textbf{Inference Mode 2: Manual Box UI.} A lightweight GUI allows the user to draw a bounding box on the overhead image and optionally type a short action command, producing an explicit visually grounded instruction.}
    \label{fig:ui_box_drawing}
\end{figure}

\subsection{Linguistic Analysis of Training Instructions}
\label{sec:linguistic_analysis}

To better understand the richness of spatial language in our training data, we analyze the distribution of atomic spatial referring terms across all instruction texts.
We identify several categories of spatial descriptors that support text-based spatial understanding:

\paragraph{Directional terms.}
Common directional references include \textit{left}, \textit{right}, \textit{top}, \textit{bottom}, \textit{upper}, \textit{lower}, \textit{front}, \textit{back}, \textit{near}, and \textit{far}.
These terms establish relative spatial relationships within the workspace.

\paragraph{Grid-based positional terms.}
For structured layouts such as egg trays and grids, instructions frequently use expressions like \textit{first row}, \textit{second row}, \textit{third row}, \textit{leftmost column}, \textit{middle column}, \textit{rightmost column}, and combinations such as \textit{second row, middle column}.

\paragraph{Relative and ordinal terms.}
Instructions also employ \textit{center}, \textit{middle}, \textit{corner}, \textit{edge}, \textit{side}, \textit{next to}, \textit{beside}, \textit{between}, \textit{closest}, and \textit{farthest} to describe object positions relative to landmarks or other objects.

\paragraph{Container and region descriptors.}
For placement tasks, terms like \textit{in the bowl}, \textit{on the tray}, \textit{inside the bin}, \textit{into the container}, and \textit{at the target location} specify destination regions.

\paragraph{Representative examples.}
The following instructions illustrate typical spatial language patterns in our dataset:
\begin{itemize}
    \item \textit{Pick the red block in the second row, middle column.}
    \item \textit{Place the object into the blue bowl on the left side.}
    \item \textit{Grasp the nearest green cube.}
    \item \textit{Move the item to the top-right corner of the tray.}
    \item \textit{Pick the object between the two containers.}
    \item \textit{Place it in the center of the workspace.}
\end{itemize}

Our training dataset contains many spatial expressions that describe relative positions, directions, and grid-like structures.
These expressions are sufficient in principle to provide unique referring descriptions for most objects and locations.
However, certain real-world scenes remain difficult for a language-only VLA to interpret with precision.
When the robot must place an object on an unstructured tabletop, or select a target in a densely cluttered scene, the spatial description cannot be specified with full clarity through language alone.
Bounding boxes provide a direct visual reference that removes the remaining ambiguity.

\subsection{Test Scenarios and Evaluation Prompts}
\label{sec:test_scenarios}

We evaluate Point-VLA on six real-world manipulation tasks that span a wide range of everyday household objects, including bottles, trays, eggs, stationery, irregular clay-like shapes, toys, and miscellaneous common items.
These scenarios capture the two major challenges faced by language-only VLA models: (1) inexpressible references, where text cannot uniquely identify the target, and (2) limited generalization, where objects or layouts differ from those seen during training.

\subsection{Detailed Text-Only Referring Instructions}

To ensure that the text-only baseline receives sufficiently informative and unambiguous language instructions, we construct for each evaluation scenario a minimal unambiguous referring expression.
These expressions contain exactly the spatial detail necessary to uniquely specify the target object or location, while remaining as concise as possible.
All instructions use only linguistic cues; no explicit visual grounding information is provided.

\subsubsection*{Irregular-shape picking}
\begin{prompt}
Pick the purple object in the front-left region of the workspace.
\end{prompt}

\subsubsection*{Cluttered picking}
\begin{prompt}
Pick the bottle on the far right of the cluster.
\end{prompt}

\subsubsection*{Plain placement}
\begin{prompt}
Place the object at the empty location in the upper-right area of the tabletop.
\end{prompt}

\subsubsection*{OOD object picking}
\begin{prompt}
Pick the black rectangular object in the lower-left area.
\end{prompt}

\subsubsection*{Egg-slot picking}
\begin{prompt}
Pick the egg in the right tray, row 2, column 3.
\end{prompt}

\subsubsection*{Egg-slot placement}
\begin{prompt}
Place the egg into the empty slot in the right tray, row 1, column 3.
\end{prompt}